\documentclass[11pt,a4paper]{article}
\usepackage[hyperref]{emnlp2018}
\usepackage{times}
\usepackage{latexsym}
\usepackage{todonotes}

\usepackage{comment}
\usepackage{graphicx}
\usepackage{amsmath}
\usepackage{multirow}
\usepackage[utf8]{inputenc} 

\aclfinalcopy 

\title{Supervised Domain Enablement Attention for Personalized Domain Classification}
\author{
  {\bf Joo-Kyung Kim} \hspace{10mm}
  {\bf Young-Bum Kim} \hspace{10mm}\\
  Amazon Alexa\\
  {\tt \{jookyk,youngbum\}@amazon.com} 
}

\date{}

\begin{document}
\maketitle
\begin{abstract}

In large-scale domain classification for natural language understanding, leveraging each user's domain enablement information, which refers to the preferred or authenticated domains by the user, with attention mechanism has been shown to improve the overall domain classification performance.
In this paper, we propose a supervised enablement attention mechanism, which utilizes sigmoid activation for the attention weighting so that the attention can be computed with more expressive power without the weight sum constraint of softmax attention.
The attention weights are explicitly encouraged to be similar to the corresponding elements of the ground-truth's one-hot vector by supervised attention, and the attention information of the other enabled domains is leveraged through self-distillation.
By evaluating on the actual utterances from a large-scale IPDA, we show that our approach significantly improves domain classification performance.
\end{abstract}

\section{Introduction}
Due to recent advances in deep learning techniques, intelligent personal digital assistants (IPDAs) such as Amazon Alexa, Google Assistant, Microsoft Cortana, and Apple Siri have been widely used as real-life applications of natural language understanding \citep{Sarikaya2016,Sarikaya2017}.

In natural language understanding, domain classification is a task that finds the most relevant domain given an input utterance \citep{Tur2011}. For example, ``make a lion sound'' and ``find me an apple pie recipe'' should be classified as \texttt{ZooKeeper} and \texttt{AllRecipe}, respectively.
Recent IPDAs cover more than several thousands of diverse domains by including third-party developed domains such as Alexa Skills \citep{Kumar2017,YBKim2018b,JKKim2018a}, Google Actions, and Cortana Skills, which makes domain classification to be a more challenging task.

Given a large number of domains, leveraging user's enabled domain information\footnote{Enabled domain information specifically refers to preferred or authenticated domains in Amazon Alexa, but it can be extended to other information such as the list of recently used domains.} has been shown to improve the domain classification performance since enabled domains reflect the user's context in terms of domain usage \citep{YBKim2018a}. For an input utterance, \citet{YBKim2018a} use attention mechanism so that a weighted sum of the enabled domain vectors are used as an input signal as well as the utterance vector. The enabled domain vectors and the attention weights are automatically trained in an end-to-end fashion to be helpful for the domain classification.

\begin{figure*}[!t]
	\centering
	\includegraphics[width=0.75\textwidth]{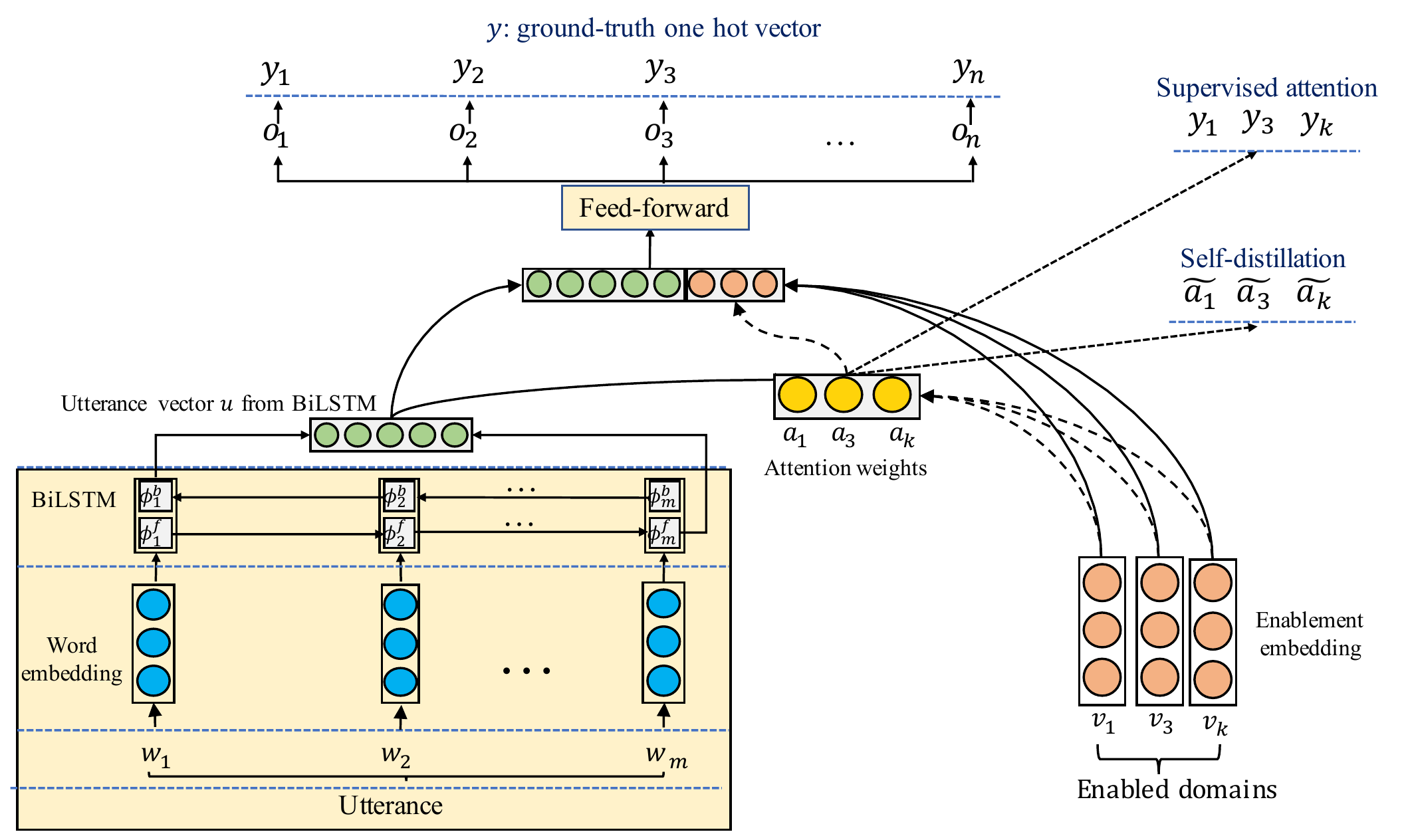}
	\caption{\small Model architecture: the input utterance is represented as a dense vector through word embedding and BiLSTM. Domain enablement vector is computed as a weighted sum of enabled domain vectors through the proposed attention mechanism. The two vectors are concatenated for the final domain prediction thorough a feed-forward neural network.}
	\label{fig:shortlister_simple}
\end{figure*}

In this paper, we propose a supervised enablement attention mechanism for more effective attention on the enabled domains. First, we use logistic sigmoid instead of softmax as the attention activation function to relax the constraint that the weight sum over all the enabled domains is 1 to the constraint that each attention weight is between 0 and 1 regardless of the other weights \citep{Martins2016, YKim2017}. Therefore, all the attention weights can be very low if there are no enabled domains relevant to a ground-truth so that we can disregard the irrelevant enabled domains, and multiple attention weights can have high values when multiple enabled domains are helpful for disambiguating an input utterance. Second, we encourage each attention weight to be high if the corresponding enabled domain is a ground-truth domain and if otherwise, to be low, by a supervised attention method \citep{Mi2016} so that the attention weights can be directly tuned for the downstream classification task. Third, we apply self-distillation \citep{Furlanello2018} on top of the enablement attention weights so that we can better utilize the enabled domains that are not ground-truth domains but still relevant.

Evaluating on datasets obtained from real usage in a large-scale IPDA, we show that our approach significantly improves domain classification performance by utilizing the domain enablement information effectively.

\section{Model}
Figure \ref{fig:shortlister_simple} shows the overall architecture of the proposed model.

Given an input utterance, each word of the utterance is represented as a dense vector through word embedding followed by bidirectional long short-term memory (BiLSTM) \citep{Graves2005}. Then, an utterance vector is composed by concatenating the last outputs of the forward LSTM and the backward LSTM.\footnote{We have also evaluated word vector summation, CNN \citep{YKim2014}, BiLSTM mean-pooling, and BiLSTM max-pooling \citep{Conneau2017} as alternative utterance representation methods, but they did not show better performance on our task.}

To represent the domain enablement information, we obtain a weighted sum of domain enablement vector where the weights are calculated by logistic sigmoid function on top of the multiplicative attention \citep{Luong2015} for the utterance vector and the domain enablement vectors.
The attention weight of an enabled domain $e$ is formulated as follows:
\begin{equation*}
a_e = \sigma\left(u \cdot v_e\right),
\end{equation*}
where $u$ is the utterance vector, $v_e$ is the enablement vector of\(\) enabled domain $e$, and $\sigma$ is sigmoid function.
Compared to conventional attention mechanism using softmax function, which constraints the sum of the attention weights to be 1, sigmoid attention has more expressive power, where each attention weight can be between 0 and 1 regardless of the other weights. We show that using sigmoid attention is actually more effective for improving prediction performance in Section \ref{sec:experiments}.

The utterance vector and the weighted sum of the domain enablement vectors are concatenated to represent the utterance and the domain enablement as a single vector.
Given the concatenated vector, a feed-forward neural network with a single hidden layer\footnote{We utilize scaled exponential linear units (SeLU) as the activation function for the hidden layer\citep{Klambauer2017}.} is used to predict the confidence score by logistic sigmoid function for each domain.

One issue of the proposed architecture is that the domain enablement can be trained to be a very strong signal, where one of the enabled domains would be the predicted domains regardless of the relevancy of the utterances to the predicted domains in many cases. To reduce this prediction bias, we use randomly sampled enabled domains instead of the correct enabled domains of an input utterance with 50\% probability during training so that the domain enablement is used as an auxiliary signal rather than determining signal. During inference, we always use the correct domain enablements of the given utterances.

The main loss function of our model is formulated as binary log loss between the confidence score and the ground-truth vector as follows:
\begin{equation*}
\mathcal{L}_{m} = -\sum_{i=1}^{n} y_i \log o_i + \left(1-y_i\right) \log \left(1-o_i\right),
\end{equation*}
where $n$ is the number of all domains, $o$ is an $n$-dimensional confidence score vector from the model, and $y$ is an $n$-dimensional one-hot vector whose element corresponding to the position of the ground-truth domain is set to 1.




\subsection{Supervised Enablement Attention}
\label{ssec:sup_es_att}
Attention weights are originally intended to be automatically trained in an end-to-end fashion \citep{Bahdanau2015}, but it has been shown that applying proper explicit supervision to the attention improves the downstream tasks such as machine translation given the word alignment and constituent parsing given annotations between surface words and nonterminals \citep{Mi2016, Liu2016, Kamigaito2017}.

We hypothesize that if the ground-truth domain is one of the enabled domains, the attention weight for the ground-truth domain should be high and vice versa. To apply this hypothesis in the model training as a supervised attention method, we formulate an auxiliary loss function as follows:
\begin{equation*}
\mathcal{L}_{a} = -\sum_{e \in E} y_e \log a_e + \left(1-y_e\right) \log \left(1-a_e\right),
\end{equation*}
where $E$ is a set of enabled domains and $a_e$ is the attention weight for the enabled domain $e$.

\subsection{Self-Distilled Attention}
One issue of supervised attention in Section \ref{ssec:sup_es_att} is that enabled domains that are not ground-truth domains are encouraged to have lower attention weights regardless of their relevancies to the input utterances and the ground-truth domains. Distillation methods utilize not only the ground-truth but also all the output activations of a source model so that all the prediction information from the source model can be utilized for more effective knowledge transfer between the source model and the target model \citep{Hinton2014}. Self-distillation, which trains a model leveraging the outputs of the source model with the same architecture or capacity, has been shown to improve the target model's performance with a distillation method \citep{Furlanello2018}.

We use a variant of self-distillation methods, where the model outputs at the previous epoch with the best dev set performance are used as the soft targets for the distillation,\footnote{This approach is closely related to Temporal Ensembling \citep{Laine2017}, but we just leverage the model outputs at the previous epoch rather than accumulating the outputs over multiple epochs.} so that the enabled domains that are not ground-truths can also be used for the supervised attention.
While conventional distillation methods utilize softmax activations as the target values, we show that distillation on top of sigmoid activations is also effective without loss of generality.
The loss function for the self-distillation on the attention weights is formulated as follows:
\begin{equation*}
\mathcal{L}_{d} = -\sum_{e \in E} \tilde{a_e} \log a_e + \left(1-\tilde{a_e}\right) \log \left(1-a_e\right),
\end{equation*}
where $\tilde{a_e}$ is the attention weight of the model showing the dev set performance in the previous epochs. It is formulated as:
\begin{equation*}
\tilde{a_e} = \sigma\left(\frac{u \cdot v_e}{T}\right),
\end{equation*}
where $T$ is the temperature for sufficient usage of all the attention weights as the soft target. In this work, we set $T$ to be 16, which shows the best dev set performance.

We have also evaluated soft-target regularization \citep{Aghajanyan2017}, where a weighted sum of the hard ground-truth target vector and the soft target vector is used as a single target vector, but it did not show better performance than self-distillation.

All the described loss functions are added to compose a single loss function as follows:
\begin{equation*}
\mathcal{L} = \mathcal{L}_{m} + \alpha\left\{\left(1-\beta\right)\mathcal{L}_{a} + \beta^t\mathcal{L}_{d}\right\},
\end{equation*}
where $\alpha$ is a coefficient representing the degree of supervised enablement attention and $\beta^t$ denotes the degree of the self-distillation. We set $\alpha$ to be 0.01 in this work. Following \citet{Hu2016}, $\beta^t = 1-0.95^t$, where $t$ denotes the current training epoch starting from 0 so that the hard ground-truth targets are more influential in the early epochs and the self-distillation is more utilized in the late epochs.

\begin{table*}[t]
\small
\centering
\begin{tabular}{c|l|lll|lll}
\multirow{2}{*}{Model no} & \multirow{2}{*}{Attention method} & \multicolumn{3}{c|}{Biased ground-truth inclusion} & \multicolumn{3}{c}{Unbiased ground-truth inclusion} \\
&                                  & Top1           & MRR            & Top3           & Top1    & MRR     & Top3   \\ \hline
(1)&sfm                               & 95.81          & 97.27          & 99.08          & 90.65   & 93.60   & 97.31  \\
(2)&sgmd                              & 95.98          & 97.43          & 99.19          & 91.03   & 93.92   & 97.49  \\
(3)&sgmd, spvs                        & 96.10          & 97.50          & 99.21          & 91.11   & 93.98   & 97.53  \\
(4)&sgmd, spvs, sdst                  & 96.29          & 97.65          & 99.32          & \textbf{91.33}    & \textbf{94.14} & \textbf{97.62}  \\
(5)&sfm, bias                         & 97.01          & 98.26          & 99.75          & 90.07   & 93.03   & 96.84  \\
(6)&sgmd, spvs, sdst, bias            & \textbf{97.48} & \textbf{98.51} & \textbf{99.76} & 90.58   & 93.30   & 96.73  \\ \hline
\end{tabular}
\caption{Accuracies (\%) on a test set with biased ground-truth inclusion in the enabled domains (90\%) (left) and a test set with unbiased inclusion (70\%) (right) with various enablement attention methods. \textbf{sftm}, \textbf{sgmd}, \textbf{spvs}, \textbf{sdst}, and \textbf{bias} denote softmax, sigmoid, supervised, self-distilled, and domain enablement bias, respectively.}
\label{tab:oneshot_result}
\end{table*}

\begin{table*}[t]
\small
\centering
\begin{tabular}{l|l|l}
Utterance                           & Ground-truth  & Enabled domain: {[}attention weights for model (1), (2), and (4){]}, ...                                                                                      \\ \hline
what is the price of bitcoin       & Crypto Price  & \begin{tabular}[c]{@{}l@{}}Sleep and Relaxation Sounds: {[}0.9998, 0.0004, 0.2029{]},\\ Crypto Price: {[}0.0001, 9.21e-0.6, 0.9977{]}\end{tabular} \\ \hline
find me a round trip ticket flight & Expedia       & Expedia: {[}0.0048, 5.37e-08, 0.6205{]}, KAYAK: {[}0.9952, 0.0004, 0.461{]}                                                                       \\ \hline
find my phone                      & Find My Phone & The Name Game: {[}1.0, 0.0001, 0.1677{]}                                                                                                           \\ \hline
\end{tabular}
\caption{Sample utterances correctly predicted with model (4) but not with model (1) and (2).}
\label{tab:example}
\end{table*}

\section{Experiments}
We evaluate our proposed model on domain classification leveraging enabled domains. The enabled domains can be a crucial disambiguating signal especially when there are multiple similar domains. For example, assume that the input utterance is \textit{``what's the weather''} and there are multiple weather-related domains such as \texttt{NewYorkWeather}, \texttt{AccuWeather}, and \texttt{WeatherChannel}. In this case, if \texttt{WeatherChannel} is included as an enabled domain of the current user, it is likely to be the most relevant domain to the user.

\label{sec:experiments}
\subsection{Datasets}
Following the data collection methods used in \citet{YBKim2018a}, our models are trained using utterances with explicit invocation patterns. For example, given a user's utterance, ``\textit{Ask} \texttt{\{ZooKeeper\}} \textit{to \{play peacock sound\}},'' \textit{``play peacock sound''} and \texttt{ZooKeeper} are extracted to compose a pair of the utterance and the ground-truth, respectively.
In this way, we have generated train, development, and test sets containing 4.4M, 500K, and 500K utterances, respectively. All the utterances are from the usage log of Amazon Alexa and the ground-truth of each utterance is one of 1K frequently used domains. The average number of enabled domains per utterance in the test sets is 8.47.

One issue of this collected data sets is that the ground-truth is included in the enabled domains for more than 90\% of the utterances, where the ground-truths are biased to enabled domains.\footnote{Since the data collection method leverages utterances where users already know the exact domain names, such domains are likely to be the enabled domains of the users.} For more correct and unbiased evaluation of the models on the input utterances from real live traffic, we also evaluate the models on the same sized train, development, and test sets where the utterances are sampled to set the ratio of ground-truth inclusion in enabled domains to be 70\%, which is closer to the ratio for actual input traffic.

\subsection{Results}
\label{ssec:results}
Table \ref{tab:oneshot_result} shows the accuracies of our proposed models on the two test sets. We also show mean reciprocal rank (MRR) and top-3, accuracy\footnote{Top-3 accuracy is calculated as \# (utterances one of whose top three predictions is a ground-truth) / \# (total utterances).} which is meaningful when utilizing post reranker, but we do not cover reranking issues in this paper \citep{Robichaud2014,YBKim2018b}.

From Table \ref{tab:oneshot_result}, we can first see that changing softmax attention to sigmoid attention significantly improves the performance. This means that having more expressive power for the domain enablement information by relaxing the softmax constraint is effective in terms of leveraging the domain enablement information for domain classification.
Along with sigmoid attention, supervised attention leveraging ground-truth slightly improves the performance, and supervised attention combined with self-distillation shows significant performance improvement. It demonstrates that supervised domain enablement attention leveraging ground-truth enabled domains is helpful, and utilizing attention information from other enabled domains is synergistic.

\citet{YBKim2018a}'s model also adds a domain enablement bias vector to the final output, which is helpful when the ground-truth domain is one of the enabled domains.
Such models (5) and (6) also show good performance for the test set where the ground-truth is one of the enabled domains with more than 90\% probability. However, for the unbiased test set where the ground-truth is included in the enabled domains with a smaller probability, not adding the bias vector is shown to be better overall.

Table \ref{tab:example} shows sample utterances correctly predicted with model (4) but not with model (1) and (2). For the first two utterances, the ground-truths are included in the enabled domains, but there were only hundreds or fewer training instances whose ground-truths are \texttt{CryptoPrice} or \texttt{Expedia}. In these cases, we can see that model (1) attends to unrelated domains, model (2) attends to none of the enabled domains, but model (4), which uses supervised attention, is shown to attend to the ground-truth even without many training examples.
``\textit{find my phone}'' has a single enabled domain which is not a ground-truth. In this case, model (1) still fully attends to the unrelated domain because of softmax attention while model (2) and (4) do not highly attend to it so that the unrelated enabled domain is not impactive.

\subsection{Implementation Details}
The word vectors are initialized with off-the-shelf GloVe vectors \citep{Pennington2014}, and all the other model parameters are initialized with Xavier initialization \citep{Glorot2010}.
Each model is trained for 25 epochs and the parameters showing the best performance on the development set are chosen as the model parameters.
We use ADAM \citep{Kingma2015} for the optimization with the initial learning rate 0.0002 and the mini-batch size 128.
We use gradient clipping, where the threshold is set to 5. We use a variant of LSTM, where the input gate and the forget gate are coupled and peephole connections are used \citep{Gers2000, Greff2017}. We also use variational dropout for the LSTM regularization \citep{Gal2016}.
All the models are implemented with DyNet \citep{Neubig2017}.

\section{Conclusion}
We have introduced a novel domain enablement attention mechanism improving domain classification performance utilizing domain enablement information more effectively. The proposed attention mechanism uses sigmoid attentions for more expressive power of the attention weights, supervised attention leveraging ground-truth information for explicit guidance of the attention weight training, and self-distillation for the attention supervision leveraging enabled domains that are not ground truth domains. Evaluating on utterances from real usage in a large-scale IPDA, we have demonstrated that our proposed model significantly improves domain classification performance by better utilizing domain enablement information.

\bibliography{emnlp2018}
\bibliographystyle{acl_natbib_nourl}
\end{document}